\documentclass[conference]{IEEEtran}
\IEEEoverridecommandlockouts
\usepackage[ruled,vlined]{algorithm2e}
\usepackage[table,xcdraw]{xcolor}

\usepackage{cite}
\usepackage{amsmath,amssymb,amsfonts}
\usepackage{graphicx}
\usepackage{textcomp}
\usepackage{multirow}
\usepackage{slashbox}
\usepackage{url,hyperref,lineno,micro type,subcaption}

\usepackage{booktabs}
\usepackage{floatrow}
\usepackage{caption}
\usepackage{wasysym}
\usepackage{wrapfig}

\def\BibTeX{{\rm B\kern-.05em{\sc i\kern-.025em b}\kern-.08em
    T\kern-.1667em\lower.7ex\hbox{E}\kern-.125emX}}

\begin{document}
\title{Topological Representations of Heterogeneous Learning Dynamics of Recurrent Spiking Neural Networks}

\author{\IEEEauthorblockN{Biswadeep Chakraborty}
\IEEEauthorblockA{\textit{Dept. of Electrical and Computer Engineering} \\
\textit{Georgia Institute of Technology}\\
Atlanta, USA \\
biswadeep@gatech.edu}
\and
\IEEEauthorblockN{Saibal Mukhopadhyay}
\IEEEauthorblockA{\textit{Dept. of Electrical and Computer Engineering} \\
\textit{Georgia Institute of Technology}\\
Atlanta, USA \\
saibal.mukhopadhyay@ece.gatech.edu}
}

\maketitle

\begin{abstract}
Spiking Neural Networks (SNNs) have become an essential paradigm in neuroscience and artificial intelligence, providing brain-inspired computation. Recent advances in literature have studied the network representations of deep neural networks. However, there has been little work that studies representations learned by SNNs, especially using unsupervised local learning methods like spike-timing dependent plasticity (STDP). Recent work by \cite{barannikov2021representation} has introduced a novel method to compare topological mappings of learned representations called Representation Topology Divergence (RTD). Though useful, this method is engineered particularly for feedforward deep neural networks and cannot be used for recurrent networks like Recurrent SNNs (RSNNs). This paper introduces a novel methodology to use RTD to measure the difference between distributed representations of RSNN models with different learning methods. We propose a novel reformulation of RSNNs using feedforward autoencoder networks with skip connections to help us compute the RTD for recurrent networks. Thus, we investigate the learning capabilities of RSNN trained using STDP and the role of heterogeneity in the synaptic dynamics in learning such representations. We demonstrate that heterogeneous STDP in RSNNs yield distinct representations than their homogeneous and surrogate gradient-based supervised learning counterparts. Our results provide insights into the potential of heterogeneous SNN models, aiding the development of more efficient and biologically plausible hybrid artificial intelligence systems.
\end{abstract}

\section{Introduction}

Spiking neural networks (SNNs) \cite{ponulak2011introduction} use bio-inspired neurons and synaptic connections, trainable with either unsupervised localized learning rules such as spike-timing-dependent plasticity (STDP) \cite{gerstner2002mathematical, chakraborty2023braindate, chakraborty2021characterization} or supervised backpropagation-based learning algorithms such as surrogate gradient \cite{neftci2019surrogate}. SNNs are gaining popularity as a powerful low-power alternative to deep neural networks for various machine learning tasks. SNNs process information using binary spike representation, eliminating the need for multiplication operations during inference. Recent advances in neuromorphic hardware \cite{akopyan2015truenorth}, \cite{davies2018loihi} have shown that SNNs can save energy by orders of magnitude compared to artificial neural networks (ANNs), maintaining similar performance levels. Since these networks are increasingly crucial as efficient learning methods, understanding and comparing the representations learned by different supervised and unsupervised learning models become essential. Empirical results on standard SNNs also show good performance for various tasks, including spatiotemporal data classification, \cite{lee2017deep,khoei2020sparnet},  sequence-to-sequence mapping \cite{chakraborty2023brainijcnn},\cite{zhang2020temporal}, object detection \cite{chakraborty2021fully,kim2020spiking}, and universal function approximation \cite{gelenbe1999function, IANNELLA2001933}.
 Furthermore, recent research has demonstrated that introducing heterogeneity in the neuronal dynamics \cite{perez2021neural, chakraborty2023heterogeneousfrontiers, chakraborty2023heterogeneousiclr, she2021heterogeneous, chakraborty2024sparse} can enhance the model's performance to levels akin to supervised learning models. Despite many previous works focusing on improving the SNN performance, unlike artificial neural networks, there has been little work to understand how these methods learn and the difference in the neuronal representations learned by the different learning methods. 

In this paper, we use a recently developed method to compare neural network representation using the topological mappings of the learned representations called Representation Topology Divergence (RTD) \cite{barannikov2021representation}. RTD is a sophisticated mathematical framework that allows researchers to evaluate the dissimilarities between distributed representations encoded within various SNN models. In contrast to traditional similarity metrics that concentrate on individual neurons or specific attributes, RTD provides a comprehensive perspective on the organization and distribution of information throughout the neural network. Drawing from algebraic topology and information theory, RTD is a robust instrument to probe the emergent characteristics of SNN representations and depicts geometric insights into the underlying computational dynamics.
However, current literature measures  RTD from the layers in feedforward networks and thus cannot be used for recurrent and reservoir-type networks like RSNNs. In this work, we focus primarily on recurrent spiking neural networks (RSNNs), as in these networks, information flows cyclically, and the temporal dynamics of neural activations play a crucial role in shaping the representations, unlike the feedforward networks. Hence, this poses a problem in utilizing the RTD-based methodology to understand the representations an RSNN model learned. Unlike traditional feedforward neural networks, which often have a bottleneck layer to compress information and extract high-level features, RSNNs do not have this explicit bottleneck. This makes it difficult to access and interpret their hidden representations directly, as isolating and analyzing the specific transformations and features learned at different time steps is challenging. Despite these challenges, understanding these representations is essential for advancing our understanding of the computational processes and improving the performance and interpretability of brain-inspired RSNN models.
Thus, this paper proposes a novel method to interpret the RSNNs as a feedforward autoencoder network with skip connections. This innovative approach provides an invaluable understanding of the complex operations in an RSNN. Simulating recurrent dynamics through feedforward structures enables a more profound grasp of the RSNN's multifaceted information processing, memory creation, and learning processes. Such progress may catalyze the development of more efficient and potent reservoir computing techniques, with broad applications including time-series prediction, signal processing, and pattern recognition. Furthermore, this methodology forms a bridge between conventional feedforward deep neural networks and the recurrent connectivity of RSNNs. Researchers can harness the extensive knowledge and tools tailored for feedforward networks to study and construct more efficient RSNN models by exposing the commonalities between these two structures. This fusion can also lead to a new generation of neural network models that synergize the efficiency of feedforward architectures with the rich dynamics of recurrent networks, crafting more biologically sound and efficient AI systems. Additionally, exploring this novel representation can enhance our comprehension of how the brain processes and preserves information, thus enriching our knowledge of human cognition and brain functionality.

In this paper, we delve into some fundamental questions concerning SNNs' behavior and performance, like understanding how the topology of the representation space evolves during unsupervised versus supervised learning and identifying the significant factors affecting the performance of different learning methods leading to diverse representations of the same input. Addressing these questions can help us better understand how SNNs learn and lead to optimizing SNN architectures and learning methods while providing deeper insights into cognitive and perceptual neural substrates. The comparison of SNN representations using RTD extends beyond individual model analysis. It assists in ensemble construction and model selection, as ensemble strategies that amalgamate diverse SNN architectures frequently exceed the performance of single models. The results from this paper can guide the choice of complementary models with varied representation spaces, resulting in enhanced generalization and resilience.
The contributions of this paper are summarized as follows:

\begin{itemize}
\item We introduce a novel reformulation of an RSNN using a feedforward autoencoder network with skip connections.
\item We employ an innovative method to access the latent space representation via the bottleneck layer, which is selected by sampling the nodes with the highest betweenness centrality from the recurrent layer
\item  We use the notion of Representation Topology Divergence on the reformulated feedforward networks to compare the representations learned by RSNN models
    \item We compare the representations learned using heterogeneous versus homogeneous STDP and compare them with those learned by backpropagated RSNN models
    \item We demonstrate that the representations learned using heterogeneous RSNN and BPRSNN are distinct, especially for temporally varying inputs, indicating that they learn distinct features
\end{itemize}

\section{Background} \label{sec:II}

\subsection{Related Works: }  
\textbf{Representation Similarity Analysis (RSA)} \cite{kriegeskorte2008representational} was developed for representation comparison between two computational models. Recent works \cite{khaligh2014deep, yamins2014performance} have used RSA to find the correlation between visual cortex features and convolutional neural network features. Raghu et al. \cite{raghu2017expressive}; Morcos et al. \cite{morcos2018insights} have
studied RSA between different neural networks. However, recent work \cite{kornblith2019similarity} argues that
none of the above methods for studying RSA can yield high similarity even between two different initializations of the same architecture. They further propose CKA, which has become a powerful evaluation tool for RSA and has been successfully applied in several studies. For example, \cite{nguyen2020wide} analyzes the representation pattern in deep and wide neural networks, and \cite{raghu2021vision} studies the
representation difference between convolutional neural networks and vision transformers with CKA. Recently, Barannikov et al. \cite{barannikov2021representation} proposed a topologically-inspired geometric approach to compare neural network representations instead of statistical methods like Canonical Correlation Analysis (CCA) and Centered Kernel Alignment (CKA). They proposed that comparing two neural representations of the same objects is the same as comparing two points clouds from different spaces. 

Despite increasing interest in pursuing high-performance SNNs with surrogate gradient training, there is limited understanding of how the different training methods affect the representation learned. Investigating this question is critical since the surrogate gradient-based BPTT algorithm mimics how ANN learns and is less biological-plausible than other learning rules like STDP. Therefore, it would be intriguing to study whether surrogate gradient-based SNNs learn different representations than unsupervised STDP-based SNNs. Understanding the representation learned in SNN can also promote further research developments, e.g., designing spiking-friendly architectures \cite{chakraborty2021fully} and exploring other ways to optimize SNNs (\cite{bellec2020solution}, \cite{zhang2020temporal}, \cite{kim2020spiking} and possibly combining surrogate gradient methods with STDP learning methods to make them more robust and biologically plausible at the same time. Though many papers study the neuronal representation among different layers of feedforward ANNs, there has been very limited study for understanding the representations learned by SNNs, especially unsupervised STDP-based learning methods. Recently, Li et al. \cite{li2023uncovering} examined the internal representation of SNNs and compared it with ANNs using the popular CKA metric. However, most of their studies compared feedforward SNNs converted from ANNs with their ANN counterparts. In this paper, we develop a novel methodology to calculate the difference in representations learned by different learning methods in RSNNs. We introduce heterogeneity in the synaptic dynamics in an RSNN and compare the representations learned with homogeneous RSNN and backpropagated RSNNs.

\subsection{Heterogeneous RSNN}

\textbf{Heterogeneous LIF Neurons} We use the Leaky Integrate and Fire (LIF) neuron model in all our simulations. In this model, the membrane potential of the $i$-th neuron $u_{i}(t)$ varies over time as:
\begin{equation}
\tau_{m} \frac{d v_{i}(t)}{d t} =-\left(v_{i}(t)-v_{rest}\right)+I_{i}(t)
\label{eq:lif}
\end{equation}
where $\tau_{\mathrm{m}}$ is the membrane time constant, $v_{rest}$ is the resting potential and $I_{\mathrm{i}}$ is the input current. When the membrane potential reaches the threshold value $v_{\text {th }}$ a spike is emitted, $v_{i}(t)$ resets to the reset potential $v_{\mathrm{r}}$ and then enters a refractory period where the neuron cannot spike. Spikes emitted by the $j$ th neuron at a finite set of times $\left\{t_{j}\right\}$ can be formalized as a spike train $\displaystyle S_{i}(t)=\sum \delta\left(t-t_{i}\right)$.

Let the recurrent layer of an RSNN be $\mathcal{R}$. We incorporate heterogeneity in the LIF neurons by using different membrane time constants $\tau_{m,i}$ and threshold voltages $v_{th,i}$ for each LIF neuron $i$ in $\mathcal{R}$. This gives a distribution of the LIF neurons' time constants and threshold voltages in $\mathcal{R}$.

\textbf{Heterogeneous STDP: } The STDP rule for updating a synaptic weight ($\Delta w$) is defined by \cite{pool2011spike}:
\begin{align}
  \Delta w(\Delta t)=\left\{\begin{array}{l}
A_{+}(w) e^{-\frac{|\Delta t|}{\tau_{+}}} \text { if } \Delta t \geq 0 \nonumber \\
-A_{-}(w) e^{-\frac{|\Delta t|}{\tau_{-}}} \text { if } \Delta t<0
\end{array}\right. \nonumber \\ 
s.t., A_{+}(w)=\eta_{+}\left(w_{\max }-w\right), A_{-}(w)=\eta_{-}\left(w-w_{\min }\right) \nonumber
\end{align}
where $\Delta t=t_{\text {post }}-t_{\text {pre }}$ is the time difference between the post-synaptic spike and pre-synaptic one, with synaptic time-constant $\tau_{\pm}$. In heterogeneous STDP, we use an ensemble of values from a distribution for $\tau_{\pm}$ and the scaling functions $\eta_{\pm}$.

While heterogeneity in neuronal parameters diversifies the network's responses to incoming stimuli, heterogeneity in STDP parameters enhances the network's ability to adapt and learn from these responses. The variability in STDP parameters allows the network to prioritize and fine-tune connections based on their functional significance. This enables the network to learn more efficiently, discriminate relevant patterns, and adapt to changing environmental conditions, making the system more robust and flexible in handling complex tasks and information processing challenges.

\begin{figure}
    \centering
    \includegraphics[width=\columnwidth]{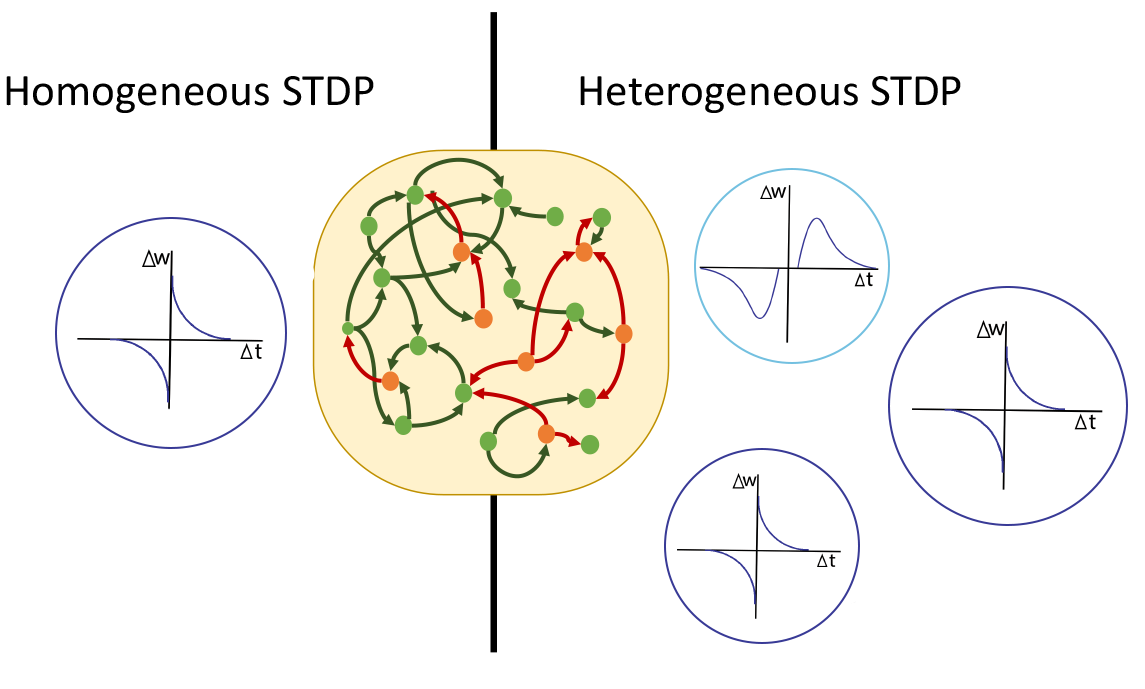}
    \caption{Figure showing the heterogeneity in the synaptic parameters}
    \label{fig:het_fig}
\end{figure}

\subsection{Similarity in Representations}
\textbf{Representation Topology Divergence (RTD)} is a mathematical framework used to quantify the dissimilarities between representations learned by complex systems. RTD leverages concepts from algebraic topology and information theory to analyze how information is organized and distributed within the high-dimensional representation spaces of these systems. RTD provides valuable insights into how different models or neural networks encode and process information by measuring the topological features, such as clustering, connectivity, and manifold structure. This analysis enables researchers to compare and understand the emergent properties of representations, identify distinctive patterns, and ultimately improve the design and performance of computational models. Moreover, RTD plays a crucial role in bridging the gap between neuroscience and artificial intelligence, fostering a deeper understanding of the brain's computational principles and inspiring more biologically plausible and efficient learning algorithms in machine learning.
This paper takes a topological perspective on the comparison of neural network representations. We propose RTD, which measures the dissimilarity between two point clouds of equal size with a one-to-one correspondence between points. Point clouds are allowed to lie in different ambient spaces.
The RTD score is defined as follows:
$RTD(X, Y) = \frac{1}{|X| |Y|} \sum_{x \in X} \sum_{y \in Y} d(x, y)$
where $d(x, y)$ is the distance between points $x$ and $y$ in the ambient space of $X$. The RTD score is effective in comparing neural network representations. In particular, it can distinguish between different classes of data and track the evolution of neural network representations during training.

\section{Methods} \label{sec:IV}

\subsection{General Framework} \label{sec:general}

\textbf{Model and Architecture: }
We empirically verify our analytical results using HRSNN for spatial and temporal classification tasks. Fig. \ref{fig:het_fig} shows the overall architecture of the model. The time-series data is encoded to a series of spike trains using a rate-encoding methodology. This high-dimensional spike train acts as the input to HRSNN. The output spike trains from HRSNN act as the input to a decoder and a readout layer that finally gives the classification results.

\textbf{Betweenness Centrality:} \label{sec:between}In a graph with recurrent connections, betweenness centrality measures how important a node is in enabling communication between different parts of the network. Nodes with high betweenness centrality are like "bridges" or "hubs" that create critical pathways for information flow between different regions of the graph. These nodes are situated on many shortest paths between other nodes, making them crucial for efficient communication and integration of information across the entire network.
The betweenness centrality of a node in a network is calculated based on the concept of shortest paths. It measures the extent to which a node lies on the shortest paths between pairs of other nodes in the network. The formula for betweenness centrality can be defined as follows:

For a given node v in the network, the betweenness centrality $(C_b(v))$ is calculated as the sum of the fraction of shortest paths passing through v for all pairs of nodes $(s, t)$ divided by the total number of possible pairs of nodes (excluding the node $v$ itself) $\displaystyle   C_b(v) = \sum_{s \neq v \neq t} \frac{\sigma(s, t|v)}{\sigma(s, t)} $. Here, $\sigma(s, t)$ represents the total number of shortest paths between nodes $s$ and $t$, while $\sigma(s, t|v)$ represents the number of those shortest paths that pass through node $v$.

\subsection{Dual Representation for Recurrent Layers}

\begin{figure}
    \centering
    \includegraphics[width=\columnwidth]{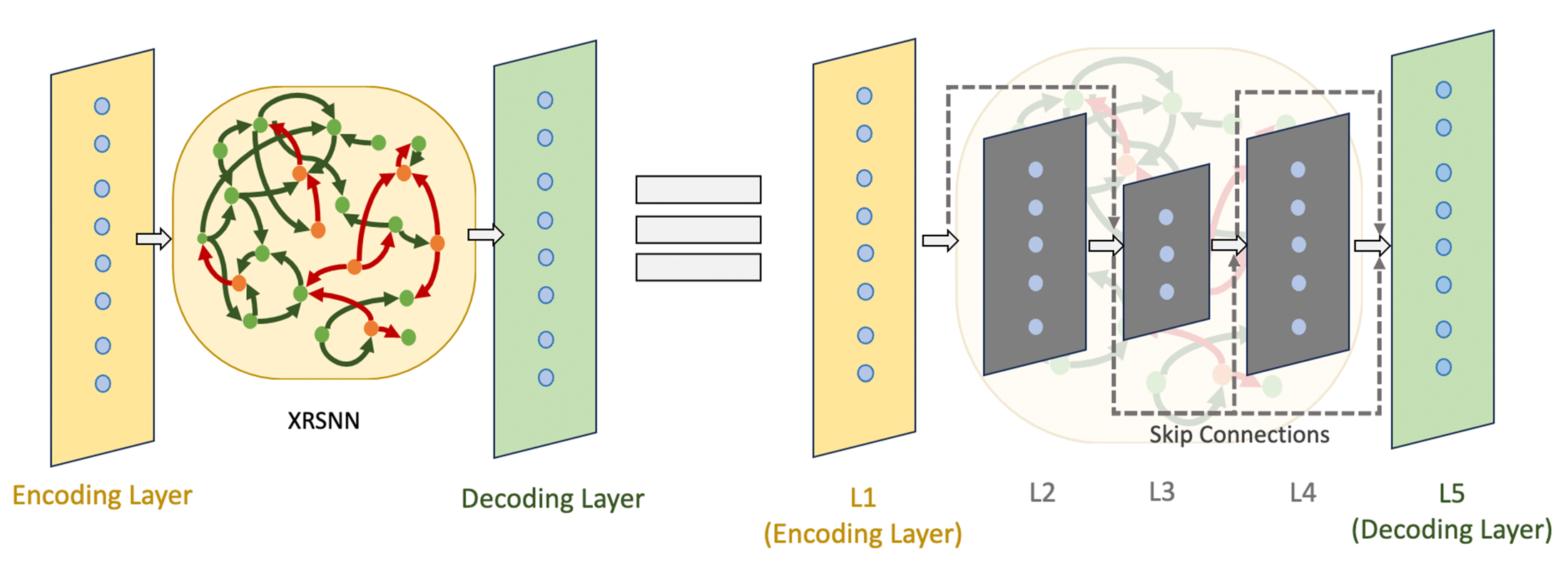}
    \caption{Figure showing how the layers are formulated in the recurrent layer}
    \label{fig:layers}
\end{figure}

We redefine the recurrent layer of an RSNN as a 5-layer feedforward autoencoder with skip connections, employing a graph-based topology where LIF neurons serve as nodes and synapses as edges, utilizing betweenness centrality to shape this representation. The layers are arranged as follows: the encoding layer ($L1$) and decoding layer ($L5$) frame the network, with neurons within these layers denoted by $n_{L1} \in L1$ and $n_{L5} \in L5$, respectively. Central to the model is the bottleneck layer ($L3$), designed for maximal information flow, flanked by intermediary layers ($L2$ and $L4$). These are strategically placed based on their proximity to $L1$, $L3$, and $L5$, quantified by the minimal number of hops between nodes, effectively reimagining the RSNN as a structured autoencoder. This compact, dual representation is illustrated in the referenced figure.
This representation is depicted in Figure \ref{fig:layers}.

More formally, Let $G $ be the graph representing the connections between nodes in the recurrent layer. Let $ L_i $ represent the $ i $-th layer of the network such that $ L_1 $ is the encoding layer, $ L_5 $ is the decoding layer, and $L_3$ is the bottleneck layer. As discussed before, we use the betweenness centrality $C_b(n)$ for each node in $G$.  Thus, we define the ordered set of neurons $M := \{n_m\} $ such that $d_{m} = \text{dist}(v_m, v_1) + \text{dist}(v_m, v_3)$ and $n_i < n_j \Leftrightarrow d_{i} < d_{j}$. Similarly, we define the ordered set of neurons $N := \{v_n\} $ such that $g_{n} = \text{dist}(v_n, v_5) + \text{dist}(v_n, v_3)$ and $v_i > v_j \Leftrightarrow g_{i} < g_{j}$. Now, we$d^{L1,L3}(n_j):= \min \|n_j - n_{L1}\| + \min \|n_j - n_{L3}$, where $\|.\|$ between two nodes $i, j$ is the number of hops required to move from node $i$ to node $j$. Thus, we define the ordered set of neurons $n^r_x \in \mathcal{X}$, where $r$ is the rank of the neuron, such that $d^{L1,L3}(n^r_x) = \|n^r_x, n_{L1}\| + \|n^r_x, n_{L3}\|$ and $n^i < n^j \Leftrightarrow d(n^i) < d(n^j)$. Similarly, we define the ordered set of neurons $n^r_y \in \mathcal{Y}$ such that $d^{L3,L5}(n^r_y) = \|n^r_y, n_{L3}\| + \|n^r_y, n_{L5}\|$ and $n^i < v^j \Leftrightarrow d(n^i) < d(n^j)$.

For the evaluations done in this paper, we use the top $10\%$ of these sets $\mathcal{X}, \mathcal{Y}$ as layer $2, 4$, respectively. In this way, we can reformulate the recurrent layer as a 5-layer autoencoder model with skip connections. It is to be noted that due to the recurrent nature of the original RSNN model, there exist connections between the layers, which can be interpreted as skip connections.  Fig. \ref{fig:layers} shows the dual representation of the RSNN.

\subsection{Comparing SNN Representations using RTD}

Representation learning from a geometric perspective relies on the manifold hypothesis, which states that real-world data found in high-dimensional space should concentrate in the vicinity of a lower-dimensional manifold. Accessing this low-dimensional manifold, denoted as $M_P$, generally requires discrete samples. A common approach involves selecting simplexes from a sample set $P$ that approximate $M_{\mathcal{P}}$ to recover the manifold. The selection process typically involves setting a threshold value $\alpha$ and choosing simplexes with edge lengths not exceeding $\alpha$. However, determining the appropriate threshold value can be challenging, so it is best to explore all thresholds.

To compare two representations of the same data, we can create corresponding graphs with distance-like weights and analyze the difference in their multiscale topology. Let $\mathcal{P}$ and $\tilde{\mathcal{P}}$ be two representations of the same data $\mathcal{V}$. Although the two embeddings belong to different ambient spaces, there exists a natural one-to-one correspondence between points in $\mathcal{P}$ and $\tilde{\mathcal{P}}$. Using a sample data set $V \subseteq \mathcal{V}$; we can create two weighted graphs $\mathcal{G}^w$ and $\mathcal{G}^{\tilde{w}}$ with the same vertex set $V$. The weights of an edge $AB$ are determined by the distances between $A$ and $B$ in $\mathcal{P}$ and $\tilde{\mathcal{P}}$.

To create a simplicial approximation of the manifold $M_{\mathcal{P}}$ at threshold $\alpha$, we select simplexes from $\mathcal{G}^w$ with edge weights not exceeding $\alpha$. Let $\mathcal{G}^{w \leq \alpha}$ represent the graph with vertex set $V$ and edges with weights not exceeding $\alpha$. To compare the simplicial approximations of $M_{\mathcal{P}}$ and $M_{\tilde{\mathcal{P}}}$ described by $\mathcal{G}^{w \leq \alpha}$ and $\mathcal{G}^{\tilde{w} \leq \alpha}$, we compare each simplicial approximation with the union of simplices formed by edges present in at least one of the two graphs. The graph $\mathcal{G}^{\min(w, \tilde{w}) \leq \alpha}$ contains an edge between vertices $A$ and $B$ if the distance between the points $A$ and $B$ is smaller than $\alpha$ in at least one of the representations $P$ or $\tilde{P}$. The set of edges of $\mathcal{G}^{\min(w, \tilde{w}) \leq \alpha}$ is the union of the sets of edges of $\mathcal{G}^{w \leq \alpha}$ and $\mathcal{G}^{\tilde{w} \leq \alpha}$. The similarity of manifolds $M_{\mathcal{P}}$ and $M_{\tilde{\mathcal{P}}}$ can be measured by the degrees of similarities of the graph $\mathcal{G}^{\min(w, \tilde{w}) \leq \alpha}$ with the graph $\mathcal{G}^{w \leq \alpha}$ and the graph $\mathcal{G}^{\tilde{w} \leq \alpha}$.

\textbf{Computing RTD for RSNNs: }To compute the RTD between distinct SNN representations, we compare their underlying topological features in the following steps:
\begin{enumerate}
    \item Convert spike trains into relevant representations capturing dynamics and patterns. Here, we utilize the population state vector, a distributed representation reflecting simultaneous population activity \cite{vyas2020computation}.
    \item Employ the Wasserstein metric described by \cite{sihn2019spike} to gauge dissimilarity between population state vectors, generating pairwise distance matrices for each RSNN.
    \item Employ dimensionality reduction (e.g., t-SNE) to project distance matrices into lower-dimensional spaces, preserving pairwise distances while reducing dimensionality.
    \item Analyze topological features by scrutinizing clustering, connectivity, and manifold structure disparities between representation spaces.
    \item Use RTD to compare topological attributes of embedded representation spaces from distinct neural networks.
\end{enumerate}

By computing RTD between two representations of different SNNs, researchers can gain valuable insights into each network's unique information processing and learning characteristics. It allows them to compare the computational dynamics of different architectures, assess model performance, and identify the strengths and weaknesses of each network in various cognitive tasks or problem domains.

\textbf{Computing Spike Distance: } In our method, the Wasserstein distance between $f$ and $g$ was adjusted for one-dimensional spike train data similar to the method developed by \cite{rubner2000earth}. We first rewrite the normalized spike trains, $f=\left\{\left(x_{1, 1} / N\right),\left(x_{2, 1} / N\right), \ldots,\left(x_{\mathrm{N}, 1} / N\right)\right\}$ and $g=$ $\left\{\left(y_{1, 1} / M\right),\left(y_{2, 1} / M\right), \ldots,\left(y_{M, 1} / M\right)\right\}$ where $x_{\mathrm{i}}$ and $y_{\mathrm{j}}$ are a sequence of spike timings. Let $d\left(x_{\mathrm{i}}, y_{\mathrm{j}}\right)$ be an absolute difference between two spike timings $x_{\mathrm{i}}$ and $y_{\mathrm{j}}$. Let $\xi_{\mathrm{ij}}$ be a flow from $x_{\mathrm{i}}$ to $y_{\mathrm{j}}$ and let $\Xi=\left[\xi_{\mathrm{ij}}\right]$ be a matrix of these flows (amount of deliveries) such that it transports $f$ to $g$ satisfying the following conditions: (1) $\xi_{\mathrm{ij}}$ is non-negative; (2) $\sum_{\mathrm{i}=1}^N \xi_{\mathrm{ij}} \leq 1 / M, \sum_{\mathrm{j}=1}^{\mathrm{M}} \xi_{\mathrm{ij}} \leq 1 / N$; and (3) $\sum_{\mathrm{i}=1}^N \sum_{\mathrm{j}=1}^{\mathrm{M}} \xi_{\mathrm{ij}}=1$. Condition 1 fixes the direction of the delivery from $i$ to $j$. Condition 2 indicates an effective delivery because it does not take back what has been delivered. Condition 3 indicates that it delivers the entire spike train. The transportation here means it makes $f$ equal to $g$ by moving parts of $f$. Then, the Wasserstein Distance between $f$ and $g$ is given by
\begin{align}
\operatorname{w}(f, g) & =\min \{\sum_{i=1}^N \sum_{j=1}^M d\left(x_i, y_j\right) \xi_{i j}: \Xi =\left[\xi_{\mathrm{ij}}\right] 
\end{align}

\section{Experiments}

\subsection{Datasets and Models}
\textbf{Datasets: } For the experiments in this paper, we primarily use CIFAR10-DVS \cite{li2017cifar10} dataset, comprising both mainly spatial components, and SHD \cite{cramer2020heidelberg} dataset comprising of mainly temporal components -  i.e., a model, to give good performance in CIFAR10-DVS needs to learn good spatial features, while the model should learn good temporal features to give good performance in SHD. This serves as a benchmark for our analyses. Due to this difference in the features of the two datasets, the representations learned from them can help us better understand and compare the different learning methodologies.

\begin{figure}
    \centering
    \includegraphics[width=\columnwidth]{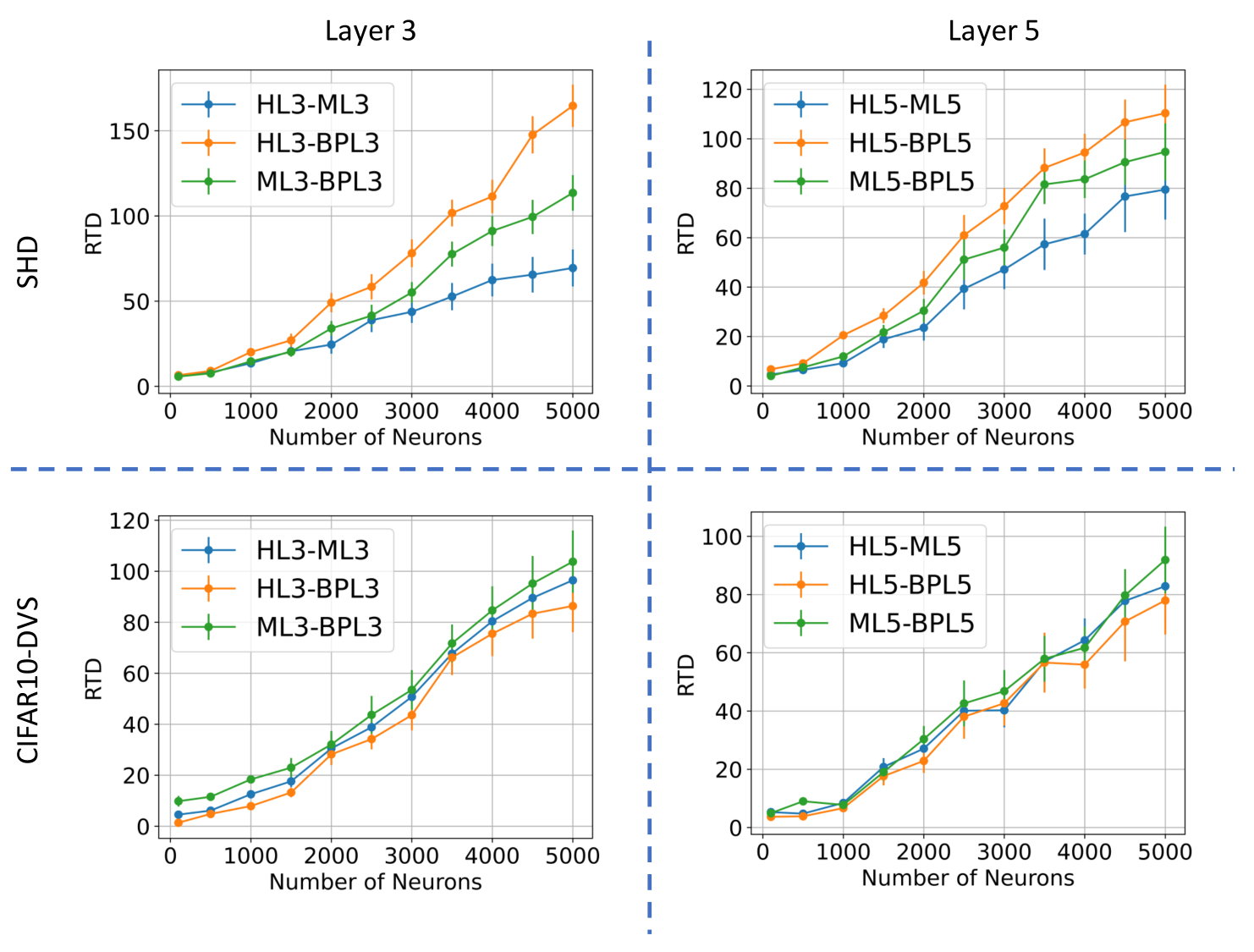}
    \caption{Layer 3, 5 RTD for Heterogeneous, Homogeneous, and BPRSNN on SHD (temporal)  and CIFAR10-DVS (spatial) task}
    \label{fig:l3}
\end{figure}

\textbf{Models:}  We use the following RSNN models for evaluation in the paper:
\begin{itemize}
    \item Homogeneous Recurrent Spiking Neural Network (MRSNN) with homogeneous STDP parameters
    \item Heterogeneous Recurrent Spiking Neural Network (HRSNN) with heterogeneity in the STDP parameters 
    \item Backpropagated Recurrent Spiking Neural Network (BPRSNN) is trained using surrogate gradient-based backpropagation. 
\end{itemize}

\textbf{Layers:} As discussed in the previous section, we can represent the recurrent layer of the RSNN as a 5-layer feedforward autoencoder network with skip connections. We label the input layer as L1, the bottleneck layer as L3, and the output layer as L5. For the three kinds of XRSNN models (X = H/M/BP), we represent the bottleneck and output layers as $XL3$ and $XL5$, respectively. This paper focuses on bottlenecks (L3) and the output layers (L5). The bottleneck layer acts as a compressed and low-dimensional representation of the input data and forces the network to capture the most significant salient features and patterns while discarding less relevant information. By condensing the input data into a reduced representation, the bottleneck layer enables efficient encoding and decoding processes and is thus paramount when comparing the learned representations of models. On the other hand, the output layer serves as the final reconstruction of the data and hence directly measures the model's performance.

\subsection{Results}
\textbf{Performance Comparison: }First, we compare the performances of the different learning models for temporal and spatio-temporal data. We evaluate the performance of the model on CIFAR10-DVS and SHD datasets and repeat the experiment 5 times to report the mean and variance of the results in Table \ref{tab:performance}. The lower triangular part of the table (marked in blue) signifies the results in CIFAR10-DVS, while the upper triangular part (marked in yellow) shows the results on the SHD dataset. We see that the performance of the heterogeneous RSNN model is comparable to that of the BPRSNN model for both the spatial and temporal datasets. We also see that heterogeneity in the synaptic dynamics helps increase performance.

\begin{table}[]
\centering
\caption{Table comparing the Accuracy for the SHD and CIFAR10-DVS classification tasks for a temporal and a spatial signal, respectively.}
\label{tab:performance}
\resizebox{0.9\columnwidth}{!}{%
\begin{tabular}{|c|c|cc|}
\hline
\multirow{2}{*}{\textbf{\begin{tabular}[c]{@{}c@{}}Learning \\ Method\end{tabular}}}           & \multirow{2}{*}{\textbf{Models}} & \multicolumn{2}{c|}{\textbf{Accuracy}}                                     \\ \cline{3-4} 
                                                                                               &                                  & \multicolumn{1}{c|}{\textit{\textbf{SHD}}} & \textit{\textbf{CIFAR10-DVS}} \\ \hline
\multirow{2}{*}{\textit{\textbf{\begin{tabular}[c]{@{}c@{}}Unsupervised\\ RSNN\end{tabular}}}} & HRSNN                            & \multicolumn{1}{c|}{80.49}                 & 69.85                         \\ \cline{2-4} 
                                                                                               & MRSNN                            & \multicolumn{1}{c|}{78.87}                 & 65.43                         \\ \hline
\textit{\textbf{\begin{tabular}[c]{@{}c@{}}Supervised\\ RSNN\end{tabular}}}                    & BPRSNN                           & \multicolumn{1}{c|}{83.54}                 & 71.68                         \\ \hline
\end{tabular}%
}
\end{table}

\begin{table}[]
\centering
\caption{Table showing the relative performance and RTD between HRSNN, MRSNN, and BPRSNN for SHD(yellow) and CIFAR10-DVS(blue) datasets}
\label{tab:rtd_acc}
\resizebox{\columnwidth}{!}{%
\begin{tabular}{|l|ll|ll|ll|}
\hline
 & \multicolumn{2}{l|}{\textbf{BPRSNN}} & \multicolumn{2}{l|}{\textbf{MRSNN}} & \multicolumn{2}{l|}{\textbf{HRSNN}} \\ \cline{2-7} 
\multirow{-2}{*}{\textbf{Models}} & \multicolumn{1}{l|}{\textit{\textbf{RTD}}} & \textit{\textbf{$\Delta$ Acc}} & \multicolumn{1}{l|}{\textit{\textbf{RTD}}} & \textit{\textbf{$\Delta$ Acc}} & \multicolumn{1}{l|}{\textit{\textbf{RTD}}} & \textit{\textbf{$\Delta$ Acc}} \\ \hline
\textbf{BPRSNN} & \multicolumn{1}{l|}{0} & 0 & \multicolumn{1}{l|}{\cellcolor[HTML]{FFFFC7}113.54} & \cellcolor[HTML]{FFFFC7}4.67 & \multicolumn{1}{l|}{\cellcolor[HTML]{FFFFC7}164.63} & \cellcolor[HTML]{FFFFC7}3.05 \\ \hline
\textbf{MRSNN} & \multicolumn{1}{l|}{\cellcolor[HTML]{CBCEFB}86.38} & \cellcolor[HTML]{CBCEFB}6.25 & \multicolumn{1}{l|}{0} & 0 & \multicolumn{1}{l|}{\cellcolor[HTML]{FFFFC7}69.47} & \cellcolor[HTML]{FFFFC7}1.62 \\ \hline
\textbf{HRSNN} & \multicolumn{1}{l|}{\cellcolor[HTML]{CBCEFB}103.76} & \cellcolor[HTML]{CBCEFB}1.83 & \multicolumn{1}{l|}{\cellcolor[HTML]{CBCEFB}96.47} & \cellcolor[HTML]{CBCEFB}4.42 & \multicolumn{1}{l|}{0} & 0 \\ \hline
\end{tabular}%
}
\end{table}

\textbf{Introduction to Representation Topology Divergence (RTD):}
The exploration of RTD between the learned representations of homogeneous and heterogeneous ESNNs sheds light on the underlying mechanisms of unsupervised learning in RSNNs. Homogeneous networks, characterized by uniform synaptic dynamics, contrast with the structural and functional diversity of heterogeneous networks. This section sets the stage by emphasizing the importance of investigating RTD to understand the impact of network diversity on learning outcomes, especially in the context of temporal and spatial data patterns.

\textbf{Comparison of RTD Across RSNN Models:}
We delve into a comparative analysis of RTD among three RSNN models, specifically focusing on the representations within the bottleneck layer (L3) and the final output layer (L5). This comparison is visualized in Figure \ref{fig:l3}, illustrating the RTD outcomes for varying neuron counts in the recurrent layer. The findings indicate minimal divergence between HRSNN and BPRSNN models for both layers, suggesting close similarities in their learned representations. However, it's important to recognize that any divergence, however minimal, is significant and indicative of differences in learning approaches.

\textbf{Detailed RTD Analysis for Different Learning Methods:}
This segment presents a detailed examination of RTD values, particularly highlighting the stark contrasts between supervised backpropagation learning (BPL) and unsupervised heterogeneous STDP learning (HL3) methods. With RTD values markedly higher in temporal datasets, the analysis underscores the superior capability of heterogeneous STDP in modeling temporal patterns. The discussion extends to comparing heterogeneous STDP-based RSNNs with backpropagation-based and homogeneous STDP-based RSNNs, revealing distinct learning representations and emphasizing the unique advantages of heterogeneity in synaptic dynamics.

\textbf{Implications of RTD Findings:}
The concluding block synthesizes the insights gained from the RTD analysis, emphasizing the broader implications for RSNN design and optimization. By highlighting the distinctive learning representations achieved through different models, this section posits that understanding RTD is crucial for advancing neural computation and machine learning. The findings not only offer a deeper comprehension of RSNN architectures but also suggest avenues for future research, particularly in enhancing RSNN performance for temporal and spatial data processing.

\textbf{Analysis of Performance-RTD Relationship:}
Here, we delve into the empirical evaluation of the performance-RTD relationship across the three RSNN models on two distinct datasets. The findings, encapsulated in Table \ref{tab:rtd_acc}, serve as the basis for this analysis. The table meticulously details the RTD values alongside the corresponding accuracy differences for each model, offering insights into how similar or divergent representations relate to actual performance outcomes. Special attention is given to the computational derivation of representational divergence, offering a formulaic perspective on interpreting the relationship between performance and RTD.

\begin{figure}
    \centering
    \includegraphics[width=\columnwidth]{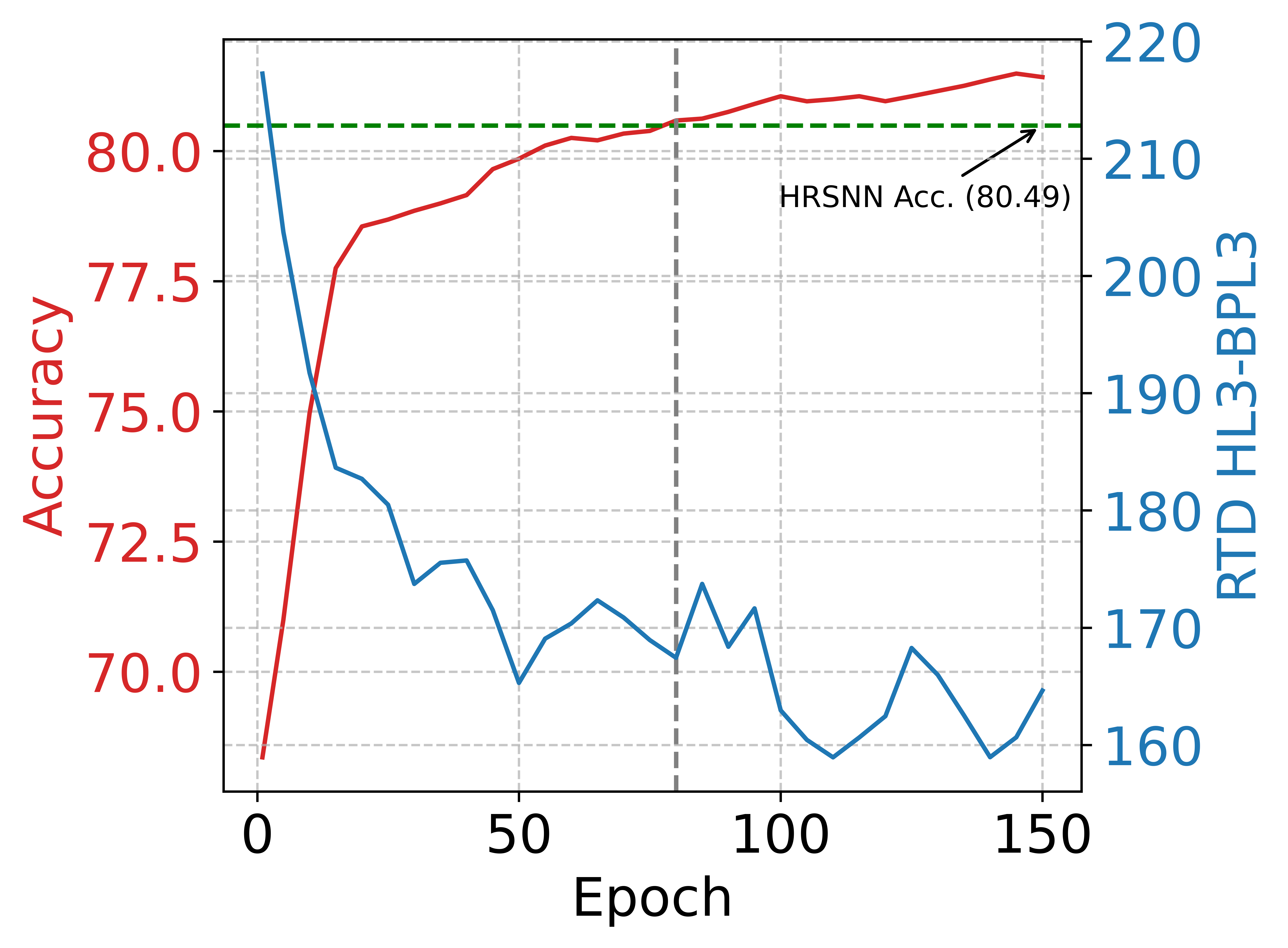}
    \caption{Figure comparing the accuracy. The left y-axis (in red) shows the accuracy of the BPRSNN model with increasing training epochs. The right y-axis shows the RTD between HL3 and BPL3 after each training epoch for the BPRSNN model with the trained HRSNN model. The green horizontal line shows the accuracy of the HRSNN model (80.49\%), and the vertical line shows the point where the accuracies of the two models are equal.}
    \label{fig:h5}
\end{figure}

\textbf{Comparative Study of Model Representations:}
In this block, the focus shifts to a comparative study of the representations between HRSNN and BPRSNN across successive training iterations. Through the lens of Figure \ref{fig:h5}, this examination scrutinizes the evolution of learned representations, especially at points of equal accuracy levels between the models. The observation of a non-negative RTD underscores the distinctiveness of the models' learning outcomes, despite similar performance metrics. This part is crucial for illustrating the practical implications of RTD in understanding the nuanced differences in learning strategies employed by RSNNs.

\textbf{Implications of Findings:}
The concluding section synthesizes the insights derived from the performance vs. RTD analysis. It reflects on the significance of identifying distinguishable learned representations between models with comparable performance, emphasizing the broader implications for neural network design and learning mechanism exploration. This block aims to contextualize the findings within the broader discourse on RSNN optimization, highlighting how such analyses contribute to a deeper understanding of the complex interplay between learning representations and model efficacy.

\section{Conclusions}

This paper proposes a novel approach to unravel recurrent spiking neural network (RSNN) learning mechanisms. By reimagining RSNNs as feedforward autoencoder networks with skip connections, we have gained insights into their information processing mechanism. This breakthrough bridges the gap between conventional feedforward architectures and the cyclic dynamics of RSNNs, offering new opportunities for more efficient reservoir computing techniques. RTD provides an insightful method for investigating the complexities of SNN representations. Using RTD, we compared the representations learned by various RSNN models, highlighting the distinctiveness of different learning methods—heterogeneous and homogeneous STDP and backpropagation-based RSNNs. Our findings underscore the potential of diverse learning approaches to capture unique features from data, advancing our comprehension of RSNN learning. This work delves into RSNNs' core, unveiling their hidden transformations through innovative reinterpretation and thorough analysis. As RSNNs continue shaping machine learning and neuromorphic computing, our work provides new insights, optimizes network architecture, and bridges the divide between biological neural systems and artificial intelligence.

\section*{Acknowledgement}
This work is supported by the Army Research Office and was accomplished under Grant Number W911NF-19-1-0447. The views and conclusions contained in this document are those of the authors and should not be interpreted as representing the official policies, either expressed or implied, of the Army Research Office or the U.S. Government.

\bibliographystyle{unsrt}
\bibliography{ref}

\end{document}